# Short-term passenger flow prediction for multi-traffic modes: A Transformer and residual network based multi-task learning method

Yongjie Yang[a], Jinlei Zhang[a]*, Lixing Yang[a]*, Xiaohong Li[b], Ziyou Gao[a]

[a] *State Key Laboratory of Rail Traffic Control and Safety, Beijing Jiaotong University, Beijing 100044, China*
[b] *School of civil engineering, Beijing Jiaotong University, Beijing 100044, China*
* *Correspondence: zhangjinlei@bjtu.edu.cn; lxyang@bjtu.edu.cn*

**Abstract**

With the prevailing of mobility as a service (MaaS), it becomes increasingly important to manage multi-traffic modes simultaneously and cooperatively. As an important component of MaaS, short-term passenger flow prediction for multi-traffic modes has thus been brought into focus. It is a challenging problem because the spatiotemporal features of multi-traffic modes are critically complex. Moreover, the passenger flows of multi-traffic modes differentiate and fluctuate significantly. To solve these problems, this paper proposes a multitask learning-based model, called Res-Transformer, for short-term inflow prediction of multi-traffic modes (subway, taxi, and bus). Each traffic mode is treated as a single task in the model. The Res-Transformer consists of two parts: (1) several modified Transformer layers comprising the conv-Transformer layer and the multi-head attention mechanism, which helps to extract the spatial and temporal features of multi-traffic modes, (2) the structure of residual network is utilized to obtain the correlations of different traffic modes and prevent gradient vanishing, gradient explosion, and overfitting. The Res-Transformer model is evaluated on two large-scale real-world datasets from Beijing, China. One is the region of a traffic hub and the other is the region of a residential area. Experiments are conducted to compare the performance of the proposed model with several baseline models. Results prove the effectiveness and robustness of the proposed method. This paper can give critical insights into the short-term inflow prediction for multi-traffic modes.

# 1 Introduction

With the rapid development of mobility as a service (MaaS), managing multi-traffic modes simultaneously and cooperatively become increasingly important. Under the environment of MaaS, people can reach their destinations by many traffic modes. The diagram of multi-traffic modes is shown in Figure 1. In some busy regions, such as traffic hubs, subway or bus stations, the pick-up or the drop-off areas of the taxi, etc., over-saturated situations often occur due to the complex passenger flow. To improve the service level and eliminate traffic congestion in these hot regions, it is important to capture the spatiotemporal distribution of the inflow for different traffic modes. As an important component of intelligent traffic systems, short-tern inflow prediction for multi-traffic modes has attracted much attention both practically and academically because it can be applied to obtaining the future passengers' volume and distribution. Accurate short-term passenger flow prediction can help operators to schedule resources more efficiently to eliminate over-saturated situations and is also helpful for passengers to plan their trips.

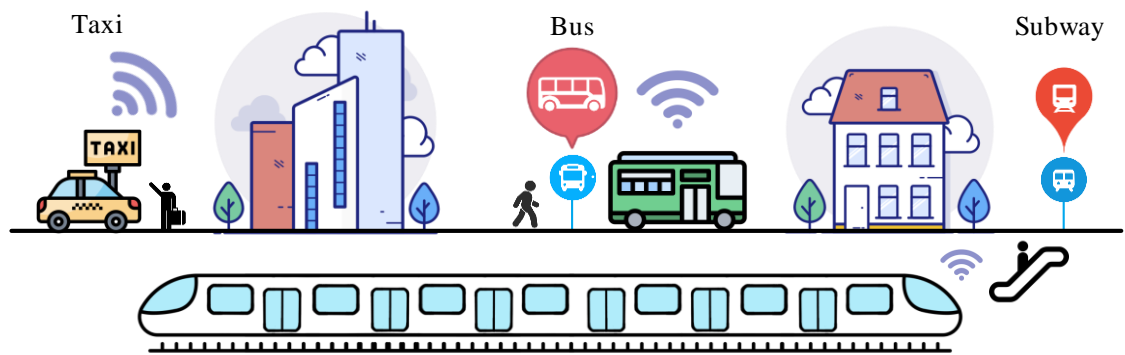

Figure 1 Diagram of multi-traffic modes

  In reality, it is significant to conduct short-term passenger flow prediction for multi-traffic modes. In terms of the short-term passenger flows of multi-traffic modes, there seem to exist different travel patterns and regularities. However, it might be predictable in some situations because people may get to or leave the same region with the same purposes by different traffic modes. For example, commuters in resident areas might take the subway, buses, or taxis on the morning of weekdays to go to work. Passengers might transfer using different traffic modes in traffic hubs. In these cases, it is feasible to conduct the short-term passenger flow prediction for multi-traffic modes. However, the existing short-term passenger flow prediction mostly aims to predict the passenger flow in a particular scene or a single traffic mode, such as subway, bus, taxi, and so forth. Studies seldom simultaneously make predictions for different traffic modes because it is challenging to capture the different travel patterns and regularities of different traffic modes and find the inside correlations of them. To overcome these problems, the main motivation of this study is thus to predict the short-term passenger flow of different traffic modes in certain regions.

  With the development of deep learning, it is feasible to make the short-term passenger flow prediction for multi-traffic modes by applying emerging deep learning techniques. Deep learning has been proved to have the ability to conduct time-series predictions. Some deep learning models are quite suitable for time series data and can extract the non-linear features inside data. The short-term passenger flow prediction for multi-traffic modes can be deemed as a time-series problem and different traffic modes show similar patterns on weekdays. However, there exists some challenges for the short-term multi-traffic modes predictions. First, even if there exist some common patterns, e.g., different traffic modes all present double peaks every weekday, it is still challenging to extract the patterns of different traffic modes simultaneously. Second, although the passenger flow of the subway in traffic hubs is quite stable, the passenger flows of taxis and buses fluctuate significantly. Deep-learning-based multitask learning techniques provide some insights for considering both the differences and similarities of multi-traffic modes. Therefore, this study tries to conduct short-term passenger flow prediction for multi-traffic modes leveraging multitask learning model.

  To tackle the aforementioned issues, this study proposes a multi-task-learning architecture called Res-Transformer, using modified Transformer layers and the structure of the residual network to predict the short-term passenger flows for multi-traffic modes. The modified Transformer layer is composed of the conv-Transformer and multi-head attention mechanism, which can better obtain the temporal features and more accurately predict the future inflow of multi-traffic modes. The structure of residual network is utilized to obatin the correlations of different traffic modes and prevent gradient vanishing, gradient explosion, and overfitting. The contributions of this paper are summarized as follows.

- To the best of our knowledge, this is the first time that a novel short-term passenger flow prediction model for multi-traffic modes is proposed.
- We propose a multi-task learning-based architecture called Res-Transformer, in which the subway, taxi, and bus are treated as different tasks. Combining the 2D CNN with the multi-head attention mechanism, we introduce a conv-Transformer layer to better extract the spatial and temporal passenger flow features of different traffic modes.
- The Res-Transformer model is tested on two large-scale real-world datasets from Beijing, China, and is compared with seven baseline models. Prediction results and five ablation analyses results prove the effectiveness and robustness of the proposed model.

The remainder of this study is organized as follows. In Section 2, we review the related work. Section 3 demonstrates the problem definition. Section 4 introduces the original Transformer model and residual network, then presents the framework overview and the proposed model. Section 5 shows the details of the experiment, including data description, the experiment platform and parameters of the model, the metrics and loss function, and the comparison with other existing models. The conclusion is summarized in Section 6.

## 2 Literature review

In this section, we classify existing short-term passenger flow prediction models into conventional statistic-based prediction models, machine-learning-based prediction models, deep-learning-based prediction models, and multitask learning-based prediction models.

### 2.1 Conventional statistic-based prediction models

Conventional models generally treat passenger flow or traffic flow as time-series data, and some statistic-based time series models are widely used for short-term traffic flow prediction, such as the autoregressive integrated average model (ARIMA), historical average model, logistic regression, Kalman filter model, and so forth (Vlahogianni et al. 2004). Ahmed and Cook (1979) were the first who applied the ARIMA model in traffic flow prediction. Cai et al. (2014) proposed a multiply ARIMA model to predict the urban railway station's inflow and outflow. The historical average model was generally used as a benchmark model. Apronti et al. (2016) applied linear and logistic regression for traffic volume estimation. Kumar (2017) proposed a Kalman filter-based model for traffic flow prediction. Nevertheless, the conventional statistic-based models cannot fully capture the spatial and temporal features of data. Therefore, the prediction accuracy is limited to some extent. In contrast, models based on machine learning and deep learning outperform these conventional models.

## 2.2 Machine-learning based prediction models

Many machine-learning models, such as the Bayesian network (Castillo et al., 2008), *k*-nearest neighbors (KNN) (Xu et al. 2020), and support vector regression (SVR) (Castro-Neto et al., 2009), show their ability in time series prediction tasks. Sun et al. (2006) used a Bayesian network to predict traffic flow. There were two highlights of this model, (1) it considered the adjacent road links and utilized them to analyze the trends of the current link, and (2) considered the data incompleteness issue when conducting traffic flow prediction. Cai et al. (2016) proposed an improved *k*-nearest neighbors (KNN) model for short-term traffic multistep prediction. The improved KNN model used a spatiotemporal state matrix to describe the traffic state of a road segment. Boukerche and Wang (2020) classified the machine-learning-based traffic flow prediction models into three types, namely, the regression models, the example-based models, and the kernel-based models. This paper throguhly analyzed the advantages and disadvantages of these machine-learning based prediction models. Although machine learning-based models perform more favorably in comparison to the conventional models, they might show their weakness when the data become massive and complex. Moreover, the machine learning-based models are generally applied to a single station or area and is hardly applied to network-level predictions or multitask learning-based predictions. With the development of deep learning, prediction models based on deep learning are more competent for the short-term passenger flow prediction compared to the machine learning-based models.

## 2.3 Deep learning based prediction models

In recent years, deep learning-based models have received great attention because of their favorable prediction performance. In the early stage, models such as the deep neural network (Liu and Chen, 2017) and the recurrent neural network (RNN) (Ma, et al., 2015) are widely used for traffic flow and passenger flow prediction. As a typical model of RNN, long short-term (LSTM) shows great potential in prediction tasks (Zhang et al., 2019). Liu et al. (2019) proposed a deep learning-based architecture called DeepPF based on the LSTM model to predict metro passenger flow. Yang et al. (2021) applied wavelet analysis to decompose the inbound passenger flow into several parts and inputted them into the LSTM model. Then, the outputs are reconstructed to form the predicted passenger flow. Lu et al. (2021) put forward a hybrid model combining ARIMA and LSTM for traffic flow prediction. Although the LSTM has the ability to fully capture the temporal features of time series data, it cannot capture the spatial features. Besides, due to the architecture of LSTM, models based on LSTM cannot be executed in parallel, which results in a long training time.

With the application of the convolutional neural network (CNN), CNN shows its ability to fully capture the spatial characters of traffic data. Zhang et al. (2021) put forward a CNN-based model called CAS-CNN for the short-term origin-destination flow prediction. The main architecture of the CAS-CNN consists of the channel-wise attention mechanism and split CNN. Liu et al. (2020) proposed a spatiotemporal ensemble net based on CNN to predict large-scale traffic states. The spatiotemporal ensemble net can combine multiple traffic state prediction models to improve the accuracy of the prediction task. Even if the CNN shows favorable performance in passenger flow or traffic flow prediction, the topological information of traffic data might be missed in the process of CNN, which may result in poor performance when conducting predictions at the network level.

Because of the ability to fully capture the spatial and temporal connection among massive stations or regions, the graph convolutional neural network (GCN) was applied to passenger flow or traffic flow predictions, and many GCN-based prediction models are widely proposed (He et al. 2022). Yu et al. (2020) proposed a novel GCN-based model, which employed both GCN and generative adversarial framework for the traffic speed prediction. The novel GCN-based model expanded the original GCN and could differentiate the intensity of connecting to neighbor roads. Tang et al. (2021) proposed a GCN-based model based on the gated recurrent unit network (GRU) and GCN to predict passenger demand at a multi-region level. Besides, this model used the Louvain algorithm to accomplish the multi-region passenger demand prediction. However, the GCN-based models can only employ several layers of GCN, once the number of layers significantly increases, it will result in poor performance. Thus, the GCN-based model cannot fully capture the high-level features of traffic data.

Many deep learning architectures combining two or more single deep learning models (RNN, CNN, GCN, et al) are proposed to overcome the shortage of a single deep learning model and improve the performance. Shi et al. (2015) first proposed a hyper model called Conv-LSTM, utilizing CNN and LSTM, for precipitation nowcasting. The Conv-LSTM model shows its advantage in capturing both spatial and temporal features. Besides, Narmadha and Vijayakumar (2021) introduced a mixed model utilizing CNN and LSTM, for spatiotemporal vehicle traffic flow prediction. As the residual network was first proposed by He et al. (2016), it has been widely used in time series prediction. Zhang et al. (2016) proposed a residual-network-based model called ST-ResNet for citywide crowd flows prediction. Zhang et al. (2021) proposed the ResLSTM, which consists of residual network, GCN, and LSTM for the short-term metro inflow prediction. Zhang et al. (2020) proposed a novel GCN and three-dimensional CNN-based model called Conv-GCN for the short-term metro inflow prediction. Li et al. (2022) put forward a novel model, called Graph-GAN, which showed a significant advantage in predicting short-term passenger flows of the subway only using the linear neural model.

All the aforementioned studies proves that there is a promising application for deep learning-based models. However, these models are formulated for a specific task or a specific traffic mode. In reality, the passenger flow might be affected by many other traffic modes. The single-task-learning-based model cannot take other traffic modes into account.

## 2.4 Multi-task learning based prediction models

Multi-task learning is capable of considering different tasks simultaneously and has been applied for many time series prediction tasks in traffic fields. Some multitask learning-based models are based on graphs. Chen et al. (2020) proposed a multitask-learning model based on GCN to predict taxi demand for a traffic road network. This paper proposed two graphs, namely, the local relationship graph which was generated according to the connectivity of road sections, and the global relationship graph, which was generated by calculating the passenger flow similarity among different road sections and treated them as two different tasks for the taxi demand prediction. Song et al. (2021) introduced a multi-task spatial-temporal graph convolutional network. In the model, the temporal and spatial features of the road sections were formed into a single graph, called the spatial-temporal graph, and using the main task and the secondary task to predict the idle time of the taxi in a specific area.

Apart from graph-based models, Zhang et al. (2020) proposed a multitask learning-based model employing GRU, for network-wide traffic speed prediction. The model treated the predictions of multiple links as multiple tasks. Huang et al. (2014) proposed a deep belief network (DBN) based on multitask learning for traffic flow prediction, and the prediction of different observation points, such as roads and stations, were treated as different tasks. Li et al. (2018) proposed a multi-task representation learning model for arrival time estimation. In this model, the main task was to predict the travel time and the auxiliary task was to predict the summaries of the path, such as travel distance, the number of links in the trip, and so forth. Mena-Yedra et al. (2018) proposed a multitask learning approach for the short-term traffic flow prediction. Zhong et al. (2017) proposed a multitask learning-based model to predict the passenger flow for a city, which treated different types of regions as different tasks. In this paper, there were three kinds of traffic modes, namely, subway, taxi, and bus. However, the passenger flow of these three traffic modes was summed up and treated as the total passenger flow of a region, and the target is to predict the total passenger flow of the region, which means they did not consider different traffic modes respectively.

Although all these aforementioned studies utilized multitask learning-based models for the prediction of traffic flows or passenger flows, most of them only focus on a single-traffic mode or treat different traffic modes as one traffic mode. With the rapid development of multi-traffic modes, conducting short-term passenger flow prediction for multi-traffic modes simultaneously become increasingly important. Therefore, it is urgent to conduct passenger flow prediction for multi-traffic modes. In this paper, we treat different traffic modes as different tasks and propose a novel multitask learning-based model to predict passenger flow for multi-traffic modes.

## 3 Problem definition

In this section, the definition of the short-term passenger flow prediction for multi-traffic modes is presented. The purpose of this study is to utilize the subway, taxi, and bus historical inflow data of a selected region to predict the inflow of the three traffic modes at the next time slot, respectively.

**Definition 1 (information of traffic modes):** We use $S, B, C$ to represent the inflow time series of the subway, bus, and taxi, respectively. The time granularity used in this study is 30 min. In this study, we only use the time span within the service time of the subway, namely from 5:00 a.m. to 11:00 p.m. Therefore, we can divide a day into a total of *18 * 60 / 30* time slots. Thus, the inflow time series can be defined as follows:

$$S = (s_{t-L}, s_{t-(L-1)}, \dots, s_{t-1}) \quad (1)$$
$$B = (b_{t-L}, b_{t-(L-1)}, \dots, b_{t-1}) \quad (2)$$
$$C = (c_{t-L}, c_{t-(L-1)}, \dots, c_{t-1}) \quad (3)$$

where $s_t$, $b_t$, $c_t$ denote the inflow of subway, bus, and taxi at the $t^{th}$ time slot, and L denotes the number of the historical time slot.

**Problem:** Given the historical inflow data of multi-traffic modes, the short-term passenger flow prediction can be formulated as a function *F(·)* that maps *L* previous inflow of the subway, bus, and taxi to the inflow of these three traffic modes at the time *t* as follows:

$$(s_{t-L}, s_{t-(L-1)}, \dots, s_{t-1}; b_{t-L}, b_{t-(L-1)}, \dots, b_{t-1}; c_{t-L}, c_{t-(L-1)}, \dots, c_{t-1}) \xrightarrow{F(\cdot)} (s_t; b_t; c_t) \quad (4)$$

For simplicity, the input of the function can be denoted as $X_{t-1} \in R^{3 \times L}$, and the output can be denoted as $Y_t \in R^{3 \times 1}$. Thus, the definition can be rewritten as follows:

$$Y_t = F(X_{t-1}) \quad (5)$$

## 4 Methodology

In this section, we firstly introduce the architecture of the Transformer and explore the details of the multi-head attention mechanism. Then, the residual network is presented. Finally, we elaborate on the details of the proposed model and introduce the framework overview.

### 4.1 Introduce of Transformer

Vaswani, A., et al. (2017) first proposed the Transformer model for Natural Language Processing (NLP). One of the highlights in the original model is the multi-head attention mechanism which is composed of multiple self-attention mechanisms. The self-attention mechanism, also called scaled dot-product attention, can be regarded as a function that can capture the correlations among different multi-traffic modes. The self-attention mechanism utilizes three parameters, namely, the inflow-related query vector *Q*, key vector *K*, and value vector *V*, to calculate the temporal dependencies among inflow data of different traffic modes. Assume the length of historical time slots is *L*, and the inflows are from the aforementioned traffic modes, the input can thus be denoted as $X \in R^{3 \times L}$. There are three weight matrixes $W\_Q \in R^{L \times d_Q}$, $W\_V \in R^{L \times d_v}$, $W\_K \in R^{L \times d_K}$, which are used to produce parameters *Q*, *K*, and *V*. For clarity, Figure 2 shows how to calculate *Q*, *K*, and *V*.

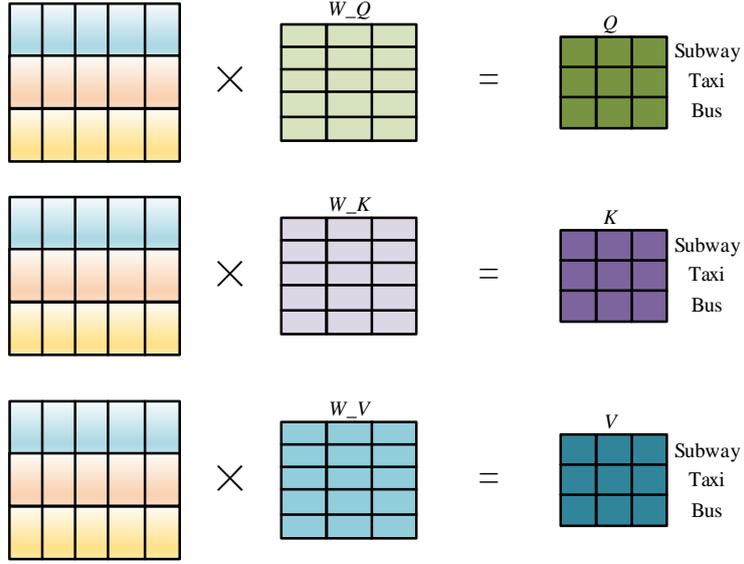

Figure 2 The calculation process of *Q*, *K*, *V*

$$Attention(Q, K, V) = Softmax(\frac{QK^T}{\sqrt{d_k}})V \quad (6)$$

The self-attention mechanism is defined as Eq. (6). Where $\sqrt{d_k}$ is used to scale the dot products, and $Softmax(\cdot)$ is the activation function, which maps the inputs into the interval [0, 1].

It is worth mentioning that, the $Softmax(QK^T/\sqrt{d_k}) \in R^{3\times3}$ represents the score matrix of the inflow data. The score matrix shows how the historical inflow data of each traffic mode affect the future inflow data. As shown in Figure 3, assume the output is $P \in R^{3\times3}$, and the first row is *P1* which represents the output of the subway. In the score matrix, the lighter the color is, the less it affects the result. As shown in the figure, the subway dominates the result *P1*, the impact factor is 0.70. Besides, compared to the bus, the influence of the taxi is more significant.

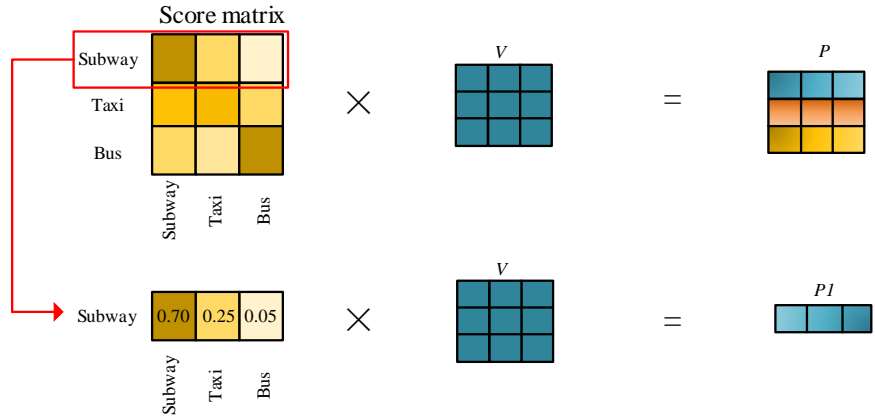

Figure 3 The mechanism of the score matrix

However, only one self-attention cannot fully capture the relations among different traffic modes. Therefore, it is necessary to gather multiple self-attention to form the multi-head attention. In the multi-head attention, each self-attention represents a head, and different heads pay attention to different features. For illustration, Figure 4 shows the architecture of the multi-head attention. Assume the shape of the input is $3 \times L$. The idea of the multi-head attention is to perform self-attention *m* times in parallel, where *m* is a hyperparameter and represents the number of self-attention layers. Then, the results of different self-attention layers are concatenated and fed into a linear layer to obtain the temporal dependency matrix of the historical inflow data.

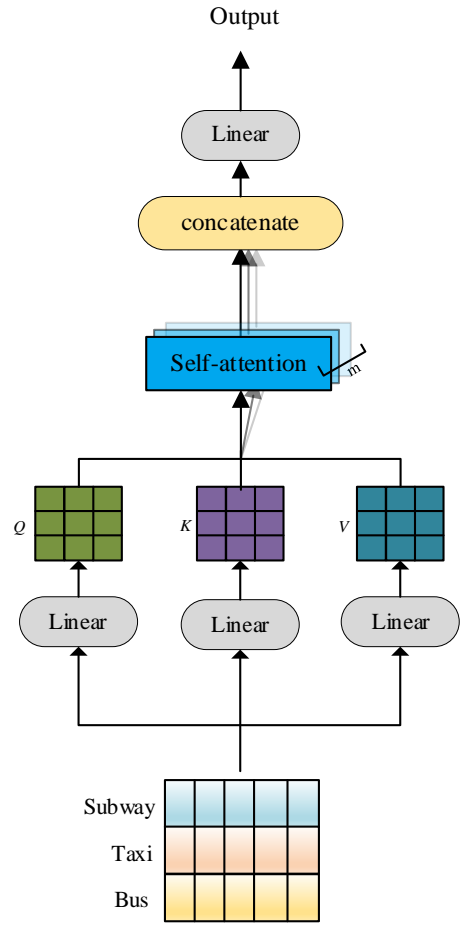

Figure 4 Diagram of multi-head attention

To show the benefits of the multi-head attention, we use the Xizhimen dataset to train the original Transformer model and obtain the score matrixes of 8 different heads. Notice that every input corresponds to a group of score matrixes, we select only one group of score matrixes for illustration. As shown in Figure 5, there are eight matrixes, each matrix consists of nine blocks, and each block indicates how the previous inflow of one traffic mode affects the future inflow. According to the color of different blocks, it is obvious that different heads pay different attention. For example, in head 2 and head 3, they seem like a cross and the color of the cross is deeper than other blocks, which indicates the passenger flow of the taxi has a majority impact on other traffic modes and is affected by all the traffic modes. Head 6 also seems like a cross, however, the color of the cross is lighter compared to the subway and bus, which is entirely different from head 2 and head 3. This example proves that a single head can only capture several features and cannot perfectly represent all features of the whole inflow data. Thus, it is worth using multi-head mechanisms to fully capture the features and relations among the inflow data of different traffic modes.

The aforementioned example shows the situation of using only one multi-head attention layer, however, just like a single head, a single multi-head attention layer can only capture limited features of the inflow data. Thus, it is necessary to set multiple layers to capture different features. The result of score matrixes obtained by four multi-head attention layers is shown in Figure 6. There are two heads in each multi-head attention layer. The heads of different layers capture entirely different information.

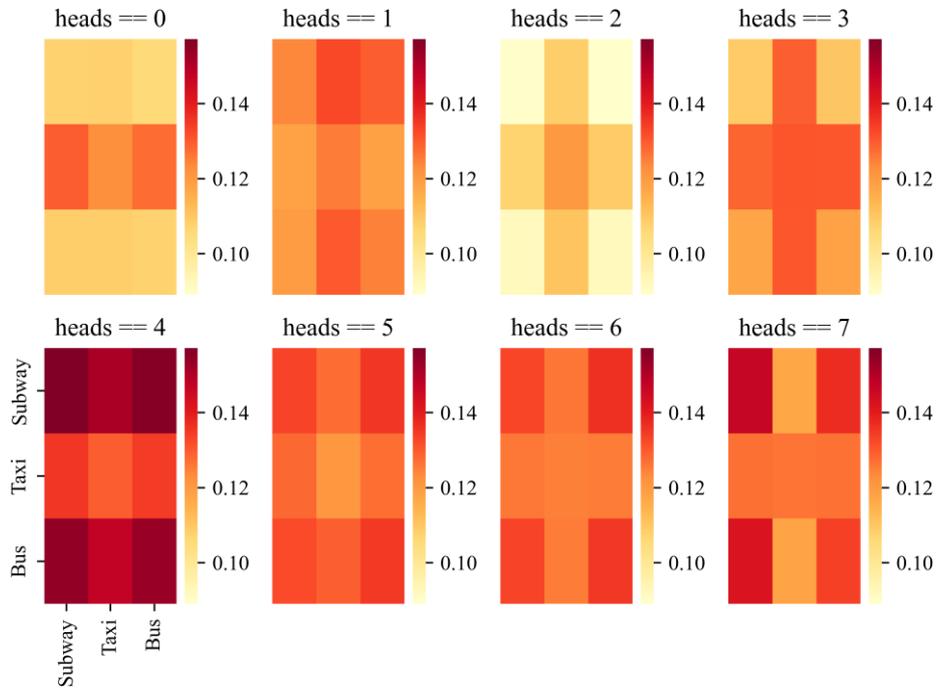

Figure 5 Score matrixes of different heads

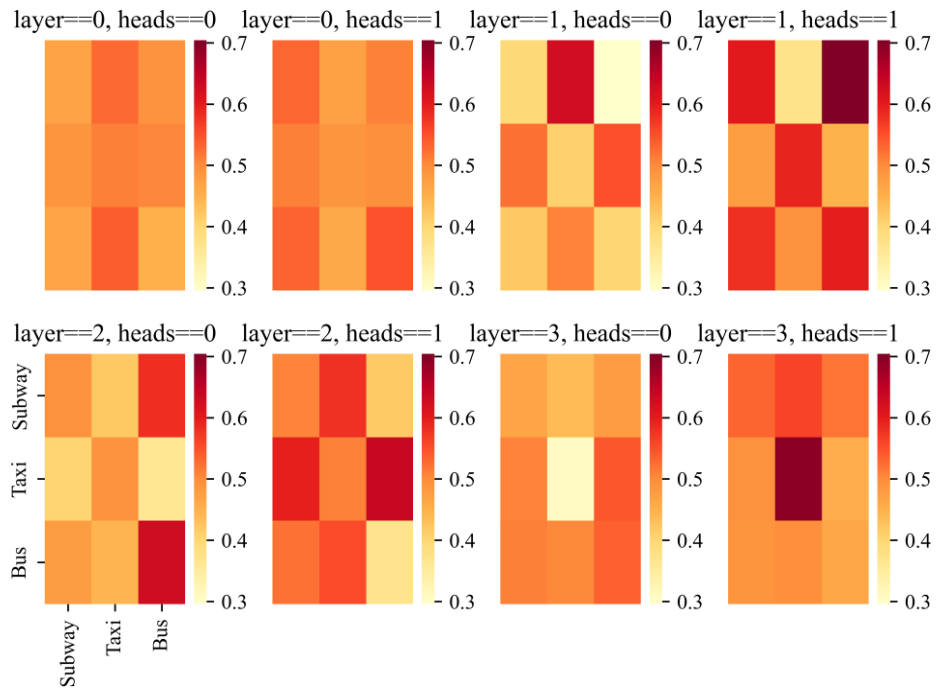

Figure 6 Score matrixes of different layers

## 4.2 Residual Network

Since the residual network was proposed (He et al. 2016), it has been widely used in many fields. For example, Zhang et al. (2016) took advantage of the residual network and proposed ST-ResNet for passenger flow prediction. The novelty of the residual network is that there is a shortcut connection, and it is designed to prevent gradient vanishing, gradient explosion, and overfitting. The architecture of the residual network is shown in Figure 7. As is shown, the value of the shortcut connection is summed up with the value of 2D convolutional layers, which helps the model to better capture the valuable information. As for the passenger flow prediction of multi-traffic modes, the 2D convolutional layers can obtain the correlations of different traffic modes at different time slots. However, some correlations might be too weak to be captured, which may result in the missing of valuable information and the difficulty of training the network. By the summation of the original input and the output of 2D convolutional layers, namely the shortcut connection, the model has the ability to fully capture the correlations among multi-traffic modes and can be trained more easily. Thus, the shortcut connection is crucial for the short-term passenger flow prediction of multi-traffic modes.

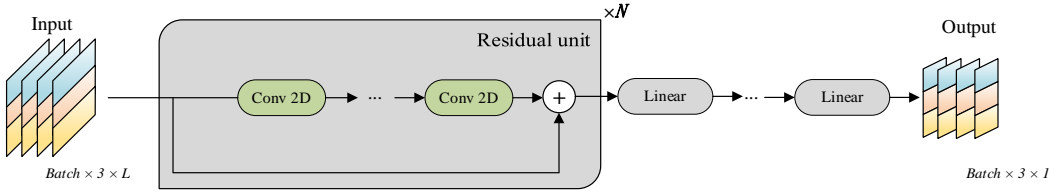

Figure 7 Architecture of residual network

## 4.3 Res-Transformer

Notice that there are two parts in the original Transformer model, namely the encoder and decoder. In this paper, we only utilize the encoder and modify it to construct the Res-Transformer. The Res-Transformer consists of two major parts, the modified transformer layers, and a shortcut connection. The architecture of the Res-Transformer is shown in Figure 8.

The Transformer plays a vital role in the Res-Transformer model because of its ability to extract the temporal dependency of inflow data. Li et al. (2019) firstly changed the architecture of the Transformer for the time series prediction. One of their contributions is that they proposed convolutional self-attention, using the convolutional layer to calculate the $Q$ and $K$ instead of using the linear layer. Enlightened by their work, we proposed the modified Transformer layer, as shown in Figure 9. The historical inflow data of three traffic modes is inputted into three different layers to calculate $Q$, $K$, and $V$, respectively. **1)** The first layer is a Conv-Transformer layer which is utilized to calculate $Q$. The Conv-Transformer layer utilizes the structure of the original Transformer and replaces the linear layers with convolutional layers. Compared to the linear layer, the convolutional layer is more suitable for processing the inflow of multi-traffic modes and can fully obtain the correlations among different traffic modes. **2)** The calculation process of $K$ and $V$ is the same as the original Transformer model. After calculating *Q*, *K*, and *V*, they are inputted into multi-head attention layers to calculate the temporal dependency of the inflow data. There are *N* modified Transformer layers and the output of the final modified Transformer layer is a matrix, we call it the information matrix. The information matrix carries the correlations of different traffic modes and each block of the information matrix represents the impact of the inflow of a traffic mode on the future inflow of the target traffic mode. Then the information matrix is fed into several convolutional layers for further processing.

Due to the meaning of each block in the information matrix, the location of the block is critical for the inflow prediction of multi-traffic modes. Thus, it is suitable to utilize the convolutional layer for further capturing the features among the blocks. After the processing of convolutional layers, the output is summed up with the original input to form the shortcut connection. Finally, the result is fed into linear layers to integrate the processed information matrix and obtain the future inflow of different traffic modes.

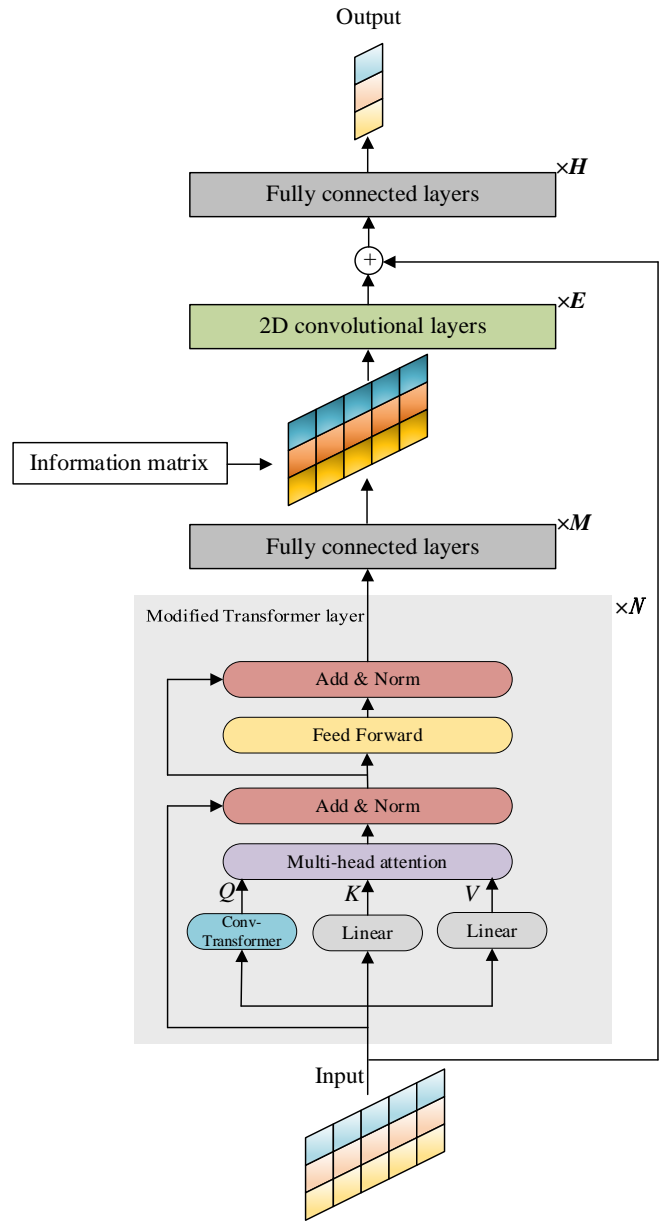

Figure 8 Architecture of Res-Transformer

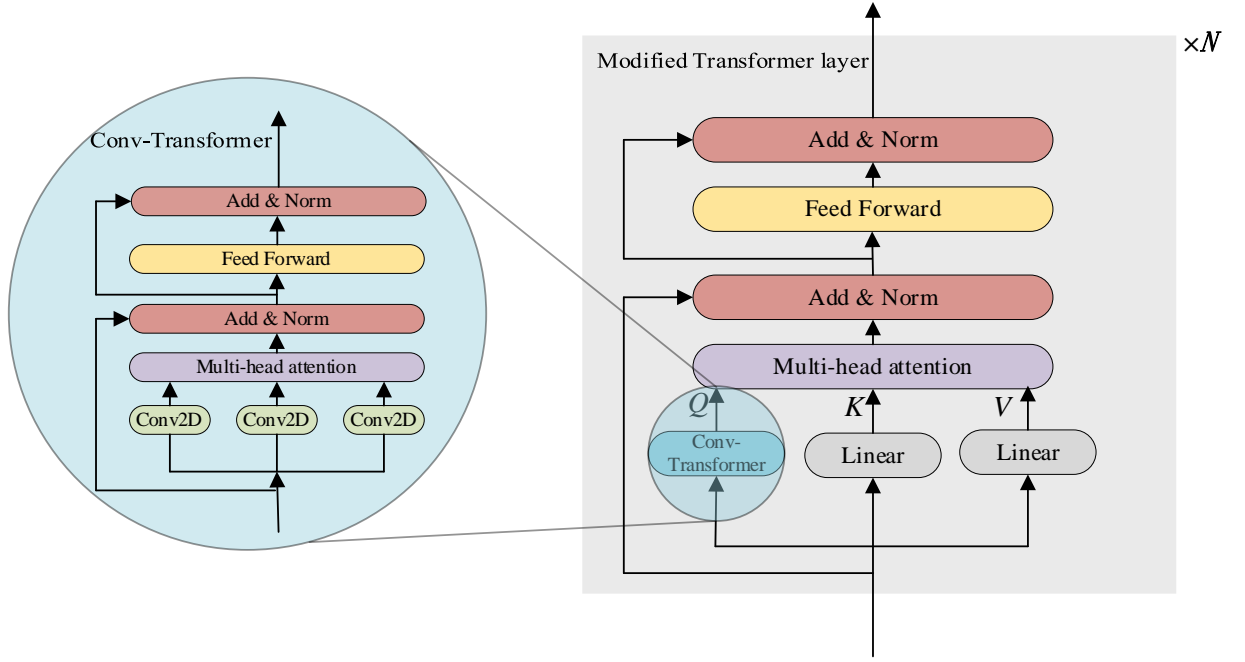

Figure 9 Details of the modified Transformer layer

The overall framework is shown in Figure 10 and it consists of two parts. Specifically, in the first part, we preprocess the data including extracting the weekday inflow data and selecting the target regions. There are a few missing records in the origin passenger flow data of subway, taxi, and bus in the selected region. Thus, we average the inflow data of the same time slot in different weeks to fill in these missing data. After processing the data, we use a sliding window to generate input $P \in R^{batch \times 3 \times L}$, where L represent the length of the historical time slots. In the second part, the historical inflow $X_{t-1}$ is fed into the Res-Transformer, which consists of several modified Transformer layers and a shortcut connection, to predict the future inflow of the three traffic modes $Y_t$ in next time slot t.

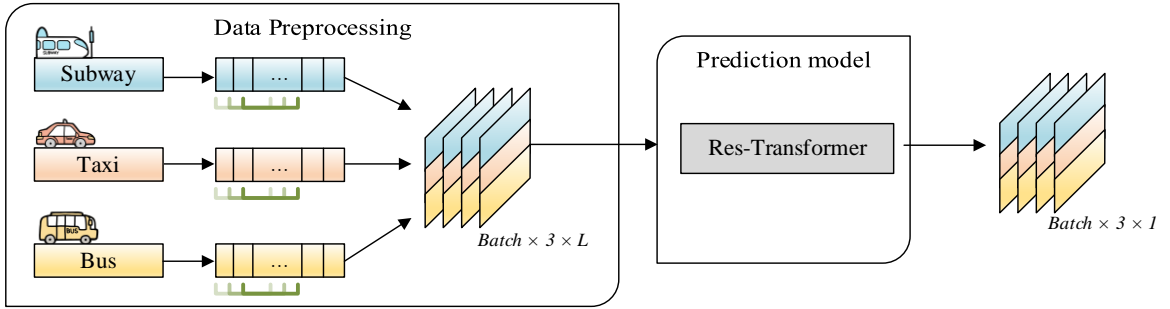

Figure 10 Framework overview

## 5  Evaluation

In this section, we first describe the datasets used in this paper. Secondly, the model configurations and evaluation metrics are represented. We conduct a series of experiments for hyperparameters tuning, to find the best performance of the Res-Transformer. Thirdly, the Res-Transformer is compared with several baseline models, namely, LSTM, CNN-1D, CNN-2D, ConvLSTM, ResNet, and Transformer. Moreover, we analyze the results of the experiments. Finally, we use the ablation analysis to validate the effectiveness of each part of the proposed model.

## 5.1 Datasets description

The experiments of this paper are conducted based on two regions of Beijing, China, namely Xizhimen (XZM) and Wangjing (WJ), which represent the traffic hub and the residential area, respectively. In each region, the dataset consists of three traffic modes, namely, subway, taxi, and bus, and we only consider the dataset ranging from February 29 to April 1, 2016 (about a month). Besides, only the weekdaya inflow data of the dataset is taken into account. Because the service times of the three traffic modes are different, in this study, we select the inflow data ranging from 5:00 a.m. to 11:00 p.m. according to the service time of the subway. The time granularity is 30 min, there are thus 36 time slots in a day.

**Subway passenger flow dataset**. We collect the AFC data of Beijing, China, which contains the records of 276 subway stations. By processing the AFC data, we obtain the subway datasets of the XZM station and WJ station, which contain weekday inflow series at the time granularity of 30min.

**Bus passenger flow dataset.** We collect the IC cards data from 2635 bus stations in Beijing. The IC cards data contains the passenger travel record, and each record contains the entry time, entry-station name, exit time, and exit-station name. First, we choose the bus stations that are within about 1,000 meters of the two selected subway stations. Then, we select the data whose entry time and exit time are within the service time of the subway and calculate the inflow series of the chosen bus stations. Since there are many bus stations in a selected region, we sum up the inflow series of the chosen bus stations and treat the summation as the total bus inflow series of the selected region.

**Taxi traffic flow dataset**. We use the TaxiBJ dataset (Zhang et al. 2016) and extract the inflow data of the two selected regions, namely XZM and WJ. The period of the extracted data is ranging from February 29 to April 1, 2016. More specifically, we then select the inflow data of a $3 \times 3$ block with the subway station as the center of the selected block to represent the region. The two selected regions, namely XZM and WJ, are shown in Figure 11. More details of the datasets are shown in Table 1.

Table 1 Details of datasets

| *DataSet* | *Subway* | *Bus* | *TaxiBJ* |
|---|---|---|---|
| *Location* | XZM and WJ | XZM and WJ | XZM and WJ |
| *Time span* | 2/29/2016 – 4/1/2016 | 2/29/2016 – 4/1/2016 | 2/29/2016 – 4/1/2016 |
| *Time Granularity* | 30min | 30min | 30min |

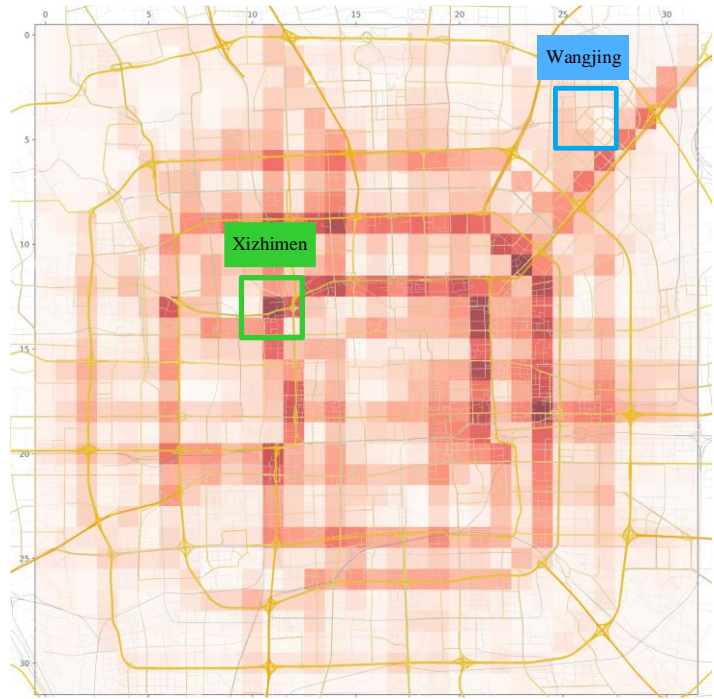

Figure 11 Two selected regions

Herein, we briefly analyze the weekday inflow data of the subway, taxi, and bus as shown in Figure 12 and Figure 13. In terms of each region, the inflow volume of the bus and subway is quite different, however, their inflow trends are similar. According to the inflow data of these two traffic modes, distinct rush hours are occurring in almost the same time slots. Compared to the rush hours of the bus and subway, the rush hours of the taxi show slight delays and the duration is longer. Overall, all of these three inflow patterns are similar, namely, they all have double peaks. In terms of the comparison between the two regions, although the inflow patterns of the two regions are quite similar, it is apparent that the inflow volume of all the traffic modes is much larger in XZM than that in WJ. In order to better capture the features of the inflow data and obtain the best results, we use max-min normalization to process the data of the two regions and map the data into the interval [-1, 1].

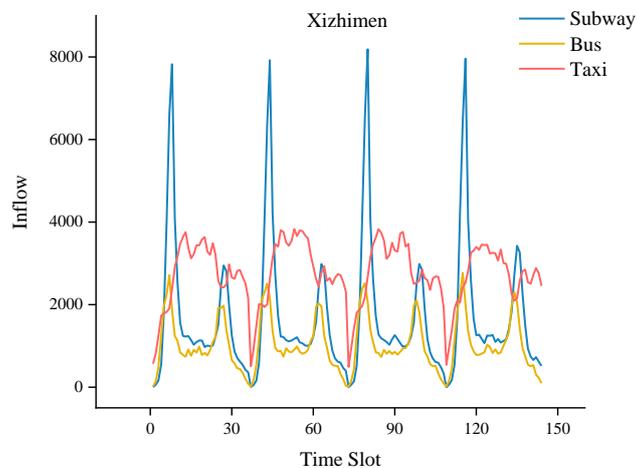

Figure 12 Inflow data of subway, taxi, bus in XZM.

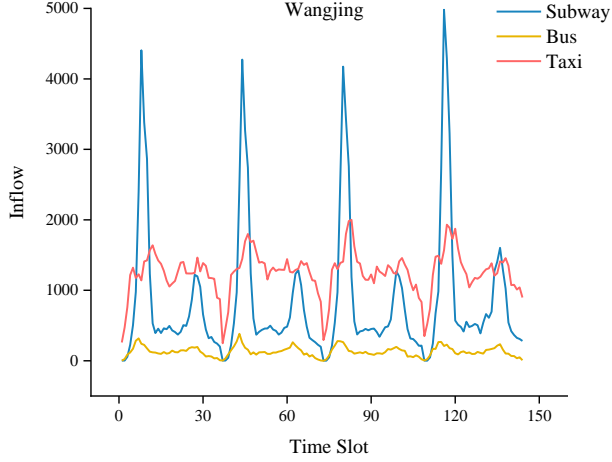

Figure 13 Inflow data of subway, taxi, bus in WJ

## 5.2 Model configurations and evaluation metrics

**Model configuration:** In the experiments, we use PyTorch to implement our model. The historical inflow data is firstly inputted into four modified Transformer layers to obtain the information matrix. The information matrix is fed into two fully connected layers, in which there are 128 neurons in the first layer, and the neuron number of the second layer is the length of historical time slots, which is treated as a hyperparameter. Then the information matrix is fed into the two convolutional layers, where the kernel size is 3×3, the stride is 1, the padding is 1, and the filters are 8. After the processing of the shortcut connection, the value is inputted into four fully connected layers with the neuron number being 128, 64, 32, and 3, respectively. Some other hyperparameters also need to be considered, including $d_q$, $d_k$, and $d_v$ in the computing process of $Q$, $K$, and $V$, length of previous time slots, number of heads in the modified Transformer layers, and batch size. The tuning of hyperparameters is demonstrated as follows.

**Hyperparameters tuning:** The value of $d_q$, $d_k$ and $d_v$ are set as the same. We set their values as (4, 8, 12, 16, 20, 24, 28, 32). The heads of the multi-head attention in the modified Transformer layers are also set as the same, and we set the heads as indicated in [2, 10]. We set the length of historical time slots as indicated in [5, 15], and the batch size is set as (2, 4, 8, 16, 32, 64, 128). In the parameters tuning process, we use the rule of control variates, which means that only one parameter will be tuned and the remaining parameters are maintained unchanged in this process until we find the best performance of the model. Take batch size as an example, firstly, we set a group of aforementioned hyperparameters randomly, and then change the value of batch size, while other parameters stay the same. Once the optimal value of batch size is found, it will not be changed anymore. Afterward, we tune the next parameter, until the four parameters are optimized and the performance of the model is the best. We use the RMSE and MAE to evaluate the performance of different parameters, and utilize the XZM dataset for the hyperparameters tuning process. The results are shown in Figure 14. As is shown, the value of $d_q$, $d_k$, and $d_v$, the length of previous time slots, the head number in the modified Transformer layers and the batch size are set as 12, 12, 4, 4, respectively.

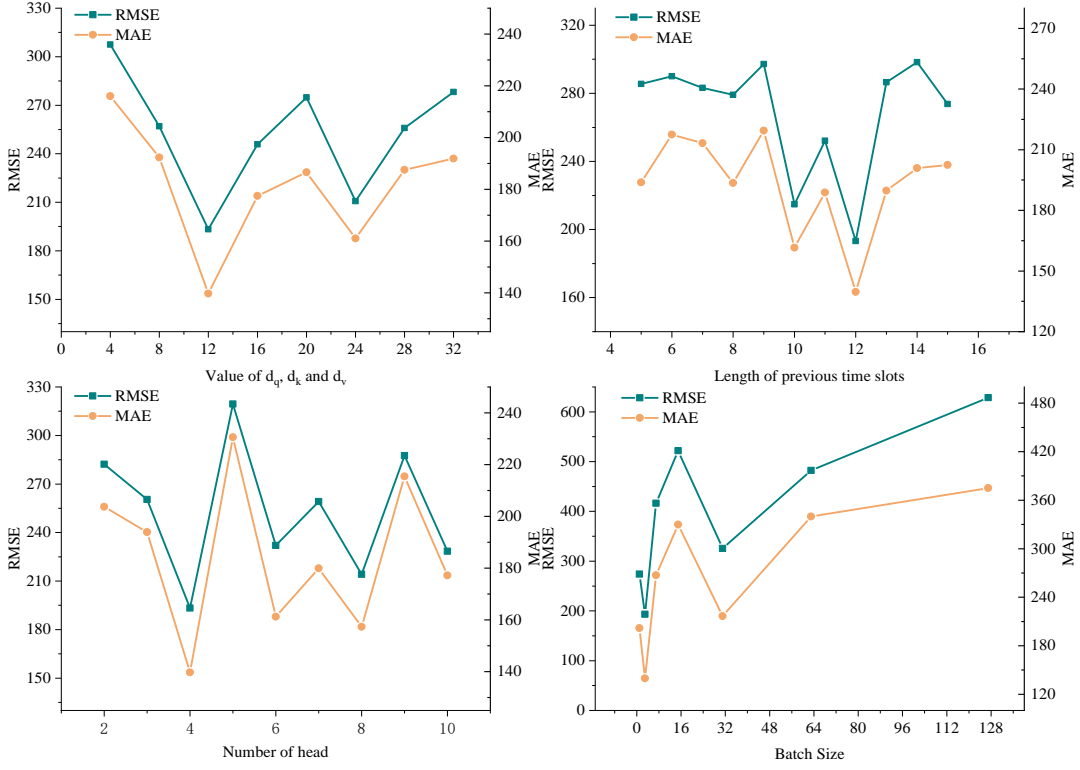

Figure 14 Hyperparameters tuning results

**Evaluation metrics:** We apply the root mean square error (RMSE), weighted mean absolute percentage error (WMAPE), and mean absolute error (MAE) to evaluate the performance of different models. Their definitions are listed below.

$$RMSE = \sqrt{\frac{1}{m}\sum_{i=1}^{m}(y_i - \hat{y}_i)^2} \qquad (7)$$

$$WMAPE = \sum_{i=1}^{m}\left(\frac{y_i}{\sum_{j=1}^{m} y_j}\left|\frac{y_i - \hat{y}_i}{y_i}\right|\right) \qquad (8)$$

$$MAE = \frac{1}{m}\sum_{i=1}^{m}|(y_i - \hat{y}_i)| \qquad (9)$$

where $y_i$ is the predicted value, $\hat{y}_i$ is the ground truth, and $m$ is the total length of the input sequence.

In addition, the mean square error (MSE) is utilized as the loss function for each traffic mode. The loss function of the Res-Transformer is the summation of the three traffic modes. The definition is given below.

$$loss = \sum_{j=1}^{3}\frac{1}{m}\sum_{i=1}^{m}(y_{ij} - \hat{y}_{ij})^2 \qquad (10)$$

where $y_{ij}$ and $\hat{y}_{ij}$ represent the predicted value and the ground truth of $j^{th}$ traffic model, and $m$ is the total length of the input sequence.

## 5.3 Baseline models

In this section, we utilize the two datasets to evaluate the performance of the Res-Transformer and compare the results with several baseline models. It is worth mentioning that, in the baseline models and the Res-Transformer, the three traffic modes are taken into account simultaneously. The details of the baseline models and the Res-Transformer are listed as follows.

**Back Propagation Neural Network (BPNN)**: As one of the most conventional machine learning models, it consists of several fully connected layers. We apply a BPNN model with three fully connected layers, and the neuron numbers of the different layers are 128, 32, and 3, respectively. We input the last 12 time slots of three traffic modes to predict the inflow of the three traffic modes in the next time slots, respectively.

**1D Convolutional Neural Network** (**CNN-1D**): We apply a CNN-1D model consisting of one 1D convolutional layer with 16 filters. The three traffic modes are treated as 3 channels and are inputted into the CNN-1D network. The kernel size is 3 with one padding and one stride. Besides, two fully connected layers are used to obtain the future inflow, and the neuron numbers are 64 and 3, respectively.

**2D Convolutional Neural Network (CNN-2D)**: In the experiment, a general CNN-2D with three fully connected layers is employed. We utilize one 2D convolutional layer with 8 filters, and the kernel size is 3×3 with one padding and one stride. Different from the CNN-1D, in the CNN-2D, three traffic modes are treated as a matrix, thus the input channel is 1. The neuron numbers of fully connected layers are 64, 32, and 3, respectively.

**Long short-term memory network (LSTM)**: As a representation of RNN, the LSTM is widely used in passenger flow prediction. We apply an LSTM model with three hidden layers, and there are 32 neurons in each layer. Besides, four fully connected layers are utilized, and the neuron numbers of them are 128, 64, 32, and 3, respectively. The inputs are the same as BPNN.

**ConvLSTM**: A hybrid model combining CNN-2D and LSTM, was proposed by Shi et al. (2015). In the experiment, a ConvLSTM model consists of three ConvLSTM layers and three fully connected layers. In terms of ConvLSTM layers, there is a 2D convolutional layer with 64 filters in each layer. In terms of fully connected layers, the neuron numbers are 64, 32, and 3, respectively. The inputs are the same as CNN-2D.

**ST-ResNet**: Zhang et al. (2016) proposed this model for citywide crowd flow prediction. There are three branches in the original model. In the experiment, we only use one branch of them. There is one residual unit and the output is fed into three fully connected layers. In the residual unit, two convolutional layers with 8 filters are utilized. The kernel size is 3×3 with one padding and one stride. The neuron numbers in the fully connected layers are 128, 64, 32, and 3, respectively. The input is the same as CNN-2D.

**Transformer**: In the Transformer, we utilize the encoder with six identical layers, and in each layer, the number of heads is 8, the values of $d_q$, $d_k$, and $d_v$ are the same and the value is 32. The output of the Transformer is inputted into four fully connected layers, and the neuron numbers are 128, 64, 32, and 3, respectively. The input is the same as BPNN.

## 5.4 Experiment results and analyses

**(1) Result analysis for all traffic modes**

We use the datasets of XZM and WJ, and the results are shown in Table 2 and Table 3. In terms of all the traffic modes, BPNN shows the poorest performance among the models in both datasets. Due to the different features of different traffic modes, the BPNN cannot efficiently identify them and might treat them as the same traffic mode. Thus, the BPNN can only capture limited nonlinear characters of inflow data, while the correlations and temporal features among different traffic modes are missing, which proves that it is not appropriate to utilize only fully connected layers. In comparison, the CNN-1D outperforms the BPNN. Although the CNN-1D can identify the different traffic modes and obtain the correlations among them, the temporal features captured by this model are limited. While the LSTM has the ability to capture the temporal features, the performance of the LSTM is thus marginally better than the CNN-1D. However, the result is yet not so good, which means it is not sufficient to only capture the temporal features or the correlations.

By taking the advantage of the LSTM and the CNN-2D, the ConvLSTM shows its ability in predicting the inflow of different modes more precisely than the former models. It might be because the initial purpose of the ConvLSTM is to predict the precipitation that the performance of a general CNN-2D model is better than the ConvLSTM. In addition, the ST-ResNet, which is also based on CNN-2D and shortcut connection, achieves better results than the CNN-2D. The performance of these models proves that the CNN-2D is the major factor that improves the performance of the model. In fact, the three traffic modes can be treated as a matrix. With the 2D convolution operation, the features of different traffic modes can be taken into account simultaneously. Thus, the model can capture the temporal features and the correlations among multi-traffic modes. With the extracted information, the model can capture the similarity of the inflow patterns among multi-traffic modes, meanwhile identifying the difference among them. However, the temporal features captured by the CNN-2D are limited, due to the structure of this model.

As is shown, the Transformer outperforms the CNN-2D, due to the multi-head attention. As mentioned in 4.1, the multi-head attention can fully capture both the temporal features and the correlations among multi-traffic modes, thus it outperforms the CNN-2D. According to the performances of the baseline models, the vital factor for the multi-traffic modes inflow prediction is to capture both temporal features and correlations thoroughly. Among the models, the CNN-2D-based models and the Transformer shows favorable performance. Hence, in order to obtain the most favorable performance for the multi-traffic modes inflow prediction, we combine the CNN-2D and the Transformer, and use the idea of the residual network, then propose the Res-Transformer. The results prove that the Res-Transformer shows the best performance in comparison to the baseline models. The Res-Transformer can capture the commonalities of the inflow patterns among multi-traffic modes precisely, as well as identify the difference among them correctly. Besides, the Res-Transformer has strong robustness in the existing models and shows promising results in both datasets. In terms of the XZM dataset, the Res-Transformer achieves the best performance with the lowest RMSE of 193.25, the lowest MAE of 139.71, and the lowest WMAPE of 8.12%, for all the traffic modes. In terms of the WJ dataset, the proposed model achieves the best performance with the lowest RMSE of 130.14, the lowest MAE of 73.81, and the lowest WMAPE of 11.36%, for all the traffic modes.

**(2) Result analysis for a single traffic mode**

In terms of the single traffic mode, as for the BPNN, CNN-1D, and LSTM, both of the BPNN and CNN-1D perform well in predicting the inflow of the bus and the taxi in both datasets, while the LSTM is better at predicting the inflow of the subway. Due to the fluctuation of the taxi and bus, the LSTM cannot fully capture their temporal features of them and accurately predict the future inflow. As for the CNN-2D-based models, namely, the ConvLSTM, the CNN-2D, and the ST-ResNet, compared to the aforementioned three models, the performance improves a lot in all the traffic modes for both datasets. Besides, in the WJ dataset, the ST-ResNet achieves the best performance in predicting the bus inflow, due to its ability to capture the spatial and temporal features when the inflow fluctuates. In fact, in both datasets, the subway inflow volume is much higher than the other traffic modes, which proves that the subway is the major traffic mode in the selected region. Especially in the WJ dataset, the inflow volume of the bus is much smaller than that of the subway and taxi. Therefore, it is more important to accurately predict the inflow of the subway and taxi, so that the results can provide more critical insights for goverments to conduct corresponding measures.

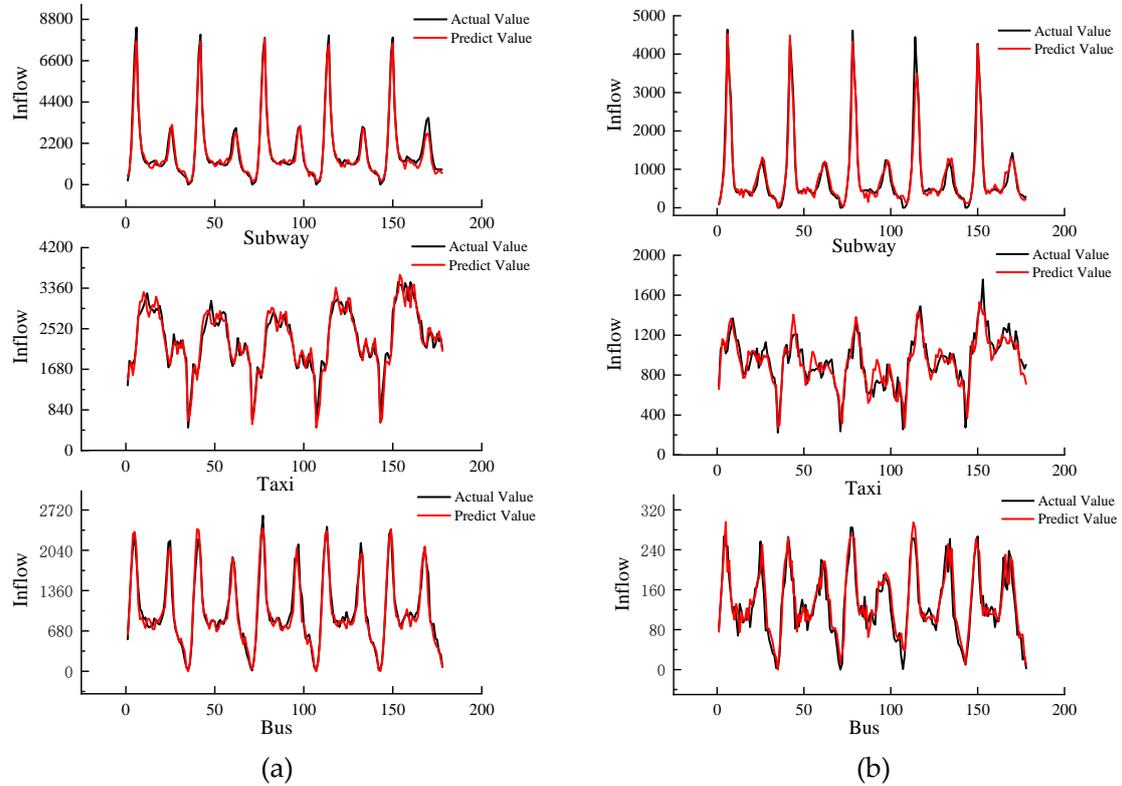

Figure 15 Prediction results of the Res-Transformer: (a) XZM dataset; (b) WJ dataset

As for the Transformer, compared to the ST-ResNet, there is a great improvement in the prediction accuracy for the subway and taxi. However, in the case of bus, the Transformer cannot fully obtain the spatiotemporal features and accurately predict the inflow. As for the Res-Transformer, it achieves the best performance in predicting the inflow of the subway and taxi, with the lowest error in both datasets. In the XZM dataset, the performance of the Res-Transformer is the best whichever the traffic modes are. In the WJ dataset, in the case of bus, the Res-Transformer performs marginally worse than the ST-ResNet, mightly due to the unstable patterns and the low volume of bus, according to Figure 15(b). However, for the purpose of better organize the transportantion resource and give an critical insights for govements, it is more important to predict the inflow of subway and taxi, especially in the WJ. The Res-Transformer is more capable for this purpose than the Transformer and the ST-ResNet.

Furthermore, the prediction results of the Res-Transformer are shown in Figure 15. As is shown, although the inflow of the three traffic modes in XZM is more stable than the inflow in WJ, the Res-Transformer performs well and the predicted values are well aligned with the actual values. In terms of the subway and bus, the Res-Transformer can precisely predict the rush hours and the off-rush hours in both datasets. In terms of the taxi, even if the rush hours are not distinct, the Res-Transformer can capture the inflow patterns and accurately predict the future inflow.

**(3) Ablation analysis**

In order to further verify the effectiveness of Res-Transformer, we conduct the ablation experiment using the XZM dataset. We change the architecture of the Res-Transformer by using the rule of control variates and construct 5 different models. We use RMSE, MAE, and WMAPE as the evaluation metrics, and the result is shown in Table 4. The five models are as follows.

**Res-Transformer (A)**: We use a fully connected layer to obtain $Q$ instead of the 2D convolutional layers in the convolutional Transformer layers.

**Res-Transformer (B)**: We remove the shortcut connection and 2D convolutional layers, and only utilize the modified Transformer layers and fully connected layers.

**Res-Transformer (C)**: We remove the 2D convolutional layers.

**Res-Transformer (D)**: We remove the shortcut connection.

**Res-Transformer (E)**: We use the encoder of the original Transformer with eight layers, instead of the modified Transformer layers, and the rest parts are the same as the Res-Transformer.

As shown in Table 4, in terms of the Res-Transformer (A) and the Res-Transformer (C) the evaluation metrics of them are much higher than the Res-Transformer, which proves the efficiency of the 2D convolutional layer. Besides, according to the performance of the Res-Transformer (A), the 2D convolutional layers inside the modified Transformer layers, which are used to produce the $Q$, are the most important among all the components. In terms of the Res-Transformer (B), the results prove that the modified Transformer layers cannot work alone. The output of the modified transformer layers, namely the information matrix that contains the features of the three traffic modes, cannot be fed into the fully connected layers directly. The information matrix needs to be further processed by the 2D convolutional layers. In terms of the Res-Transformer (D), if we drop the shortcut connection, the model performs marginally worse than the proposed model. Thus, to obtain the precise future inflow, the information matrix needs to be further processed by the 2D convolutional layers and summed up with the original input, as in our proposed model.

In terms of the Res-Transformer (E), which replaces the modified Transformer layers with the encoder of the original Transformer model, shows the best performance among the aforementioned models. Besides, compared to the performance of the Transformer in

Table 2 Model performance comparison of subway, bus, taxi in XZM

| XZM | Subway | | | Taxi | | | Bus | | | ALL | | |
|---|---|---|---|---|---|---|---|---|---|---|---|---|
| | RMSE | MAE | WMAPE | RMSE | MAE | WMAPE | RMSE | MAE | WAMPE | RMSE | MAE | WMAPE |
| BPNN | 409.51 | 302.02 | 16.39% | 249.93 | 202.28 | 8.79% | 167.40 | 129.59 | 12.70% | 293.37 | 211.30 | 12.26% |
| CNN-1D | 404.32 | 279.94 | 15.19% | 245.16 | 197.88 | 8.60% | 181.55 | 137.29 | 13.45% | 292.43 | 205.04 | 11.90% |
| LSTM | 328.56 | 228.89 | 12.52% | 291.32 | 232.62 | 10.10 | 187.25 | 143.76 | 14.09% | 275.61 | 201.76 | 11.75% |
| ConvLSTM | 347.28 | 282.49 | 15.44% | 276.98 | 238.57 | 10.37% | 148.90 | 115.66 | 11.33% | 270.49 | 212.24 | 12.36% |
| CNN-2D | 321.01 | 250.04 | 13.58% | 257.84 | 210.67 | 9.15% | 138.28 | 104.02 | 10.19% | 250.77 | 188.25 | 10.93% |
| ST-ResNet | 321.63 | 235.80 | 12.88% | 229.90 | 180.85 | 7.86% | 138.60 | 105.49 | 10.34% | 241.88 | 174.05 | 10.13% |
| Transformer | 286.89 | 215.36 | 11.78% | 222.92 | 171.51 | 7.45% | 143.81 | 111.62 | 10.94% | 225.60 | 166.16 | 9.68% |
| Res-Trans | **251.57** | **183.69** | **10.00%** | **173.70** | **136.30** | **5.92%** | **136.30** | **99.13** | **9.71%** | **193.25** | **139.71** | **8.12%** |

Table 3 Model performance comparison of subway, bus, taxi in WJ

| WJ | Subway | | | Taxi | | | Bus | | | ALL | | |
|---|---|---|---|---|---|---|---|---|---|---|---|---|
| | RMSE | MAE | WMAPE | RMSE | MAE | WMAPE | RMSE | MAE | WAMPE | RMSE | MAE | WMAPE |
| BPNN | 337.58 | 213.99 | 24.18% | 121.52 | 93.50 | 9.91% | 29.90 | 24.10 | 18.35% | 207.86 | 110.53 | 16.77% |
| CNN-1D | 324.23 | 210.14 | 23.48% | 112.72 | 87.62 | 9.28% | 31.25 | 24.00 | 18.38% | 199.00 | 107.25 | 16.16% |
| LSTM | 303.77 | 213.60 | 24.31% | 155.01 | 114.77 | 12.16% | 31.84 | 25.59 | 19.63% | 197.75 | 117.99 | 18.02% |
| ConvLSTM | 222.52 | 168.89 | 19.53% | 131.70 | 105.98 | 11.23% | 34.84 | 27.62 | 21.14% | 150.64 | 100.83 | 15.56% |
| CNN-2D | 214.34 | 161.13 | 18.41% | 141.89 | 115.31 | 12.22% | 32.83 | 25.74 | 19.64% | 149.62 | 100.72 | 15.45% |
| ST-ResNet | 212.68 | 161.13 | 18.80% | 122.06 | 96.33 | 10.20% | **27.51** | **21.74** | **16.59%** | 142.46 | 93.07 | 14.43% |
| Transformer | 200.04 | 132.30 | 15.63% | 134.55 | 104.93 | 11.12% | 28.80 | 22.87 | 17.49% | 140.19 | 95.86 | 14.68% |
| Res-Trans | **196.50** | **113.42** | **13.01%** | **106.75** | **85.54** | **9.06%** | 28.29 | 22.48 | 17.25% | **130.14** | **73.81** | **11.36%** |

Table 4 Ablation analysis results of subway, taxi, bus in XZM

| XZM | Subway | | | Taxi | | | Bus | | | ALL | | |
|---|---|---|---|---|---|---|---|---|---|---|---|---|
| | RMSE | MAE | WMAPE | RMSE | MAE | WMAPE | RMSE | MAE | WAMPE | RMSE | MAE | WMAPE |
| Res-Trans(A) | 498.00 | 408.86 | 22.08% | 213.10 | 175.38 | 7.62% | 139.58 | 113.36 | 11.11% | 322.95 | 132.53 | 13.44% |
| Res-Trans(B) | 406.81 | 306.23 | 16.68% | 318.63 | 259.69 | 11.28% | 173.59 | 132.86 | 13.02% | 314.72 | 232.93 | 13.54% |
| Res-Trans(C) | 364.49 | 286.48 | 15.69% | 225.04 | 176.20 | 7.66% | 195.15 | 149.87 | 14.69% | 271.77 | 204.19 | 11.90% |
| Res-Trans(D) | 280.11 | 215.34 | 11.80% | 251.72 | 217.83 | 9.46% | 141.44 | 104.81 | 10.27% | 235.92 | 179.33 | 10.45% |
| Res-Trans(E) | 280.42 | 202.43 | 11.09% | 185.93 | 149.65 | 6.50% | 140.80 | 110.46 | 10.82% | 210.58 | 154.18 | 8.90% |
| Res-Trans | **251.57** | **183.69** | **10.00%** | **173.70** | **136.30** | **5.92%** | **136.30** | **99.13** | **9.71%** | **193.25** | **139.71** | **8.12%** |

Table 2, if we use the architecture of the Res-Transformer with original Transformer layers, the prediction accuracy is better than the Transformer in all traffic modes, which proves the effectiveness of the proposed architecture. However, the Res-Transformer still outperforms Res-Transformer (E), which proves that the original Transformer layers cannot thoroughly capture the temporal features and correlations among multi-traffic modes and extract the valuable information matrix.

# 6  Conclusion and future work

This study proposed a novel multitask-learning architecture called Res-Transformer, which can accurately predict the future inflow of multi-traffic modes. The Res-Transformer is composed of modified Transformer layers and the residual network. The main conclusions can be summarized as follows.

- The Res-Transformer has significant advantages to obtain the correlations and the temporal features of the inflow among the selected traffic modes (subway, taxi, and bus). The proposed conv-Transformer layer is effective to capture and integrate the features of multi-traffic modes.
- According to the performance of the Res-Transformer and baseline models, the LSTM shows weakness in the situation of multi-traffic modes, and the Transformer and CNN-2D-based models are much more suitable for inflow prediction of the multi-traffic modes in the selected region.
- The results tested on two large-scale real-world datasets shows that the Res-Transformer has strong robustness and significant improvements compared to the best existing models, in terms of the XZM dataset, the Res-Transformer achieves the best performance with the lowest RMSE, MAE, and WMAPE of 193.25, 139.71, and 8.12%, for all the traffic modes, respectively. In terms of the WJ dataset, the proposed model achieves the best performance with the lowest RMSE, MAE, and WMAPE of 130.14, 73.81, and 11.36%, for all the traffic modes, respectively.

However, there are some limitations to this study. First, we only consider subway, taxi, and bus in this paper, but there are many other traffic modes to choose from apart from these three modes. Second, this study only considers two typical regions, however, we do not consider the whole traffic network, for example, the whole city. Besides, it is well known that the weather conditions will affect the passenger's choices of travel to a certain degree, but we do not consider this in our study. Also, the inflow we select is from weekday, while we do not take the weekend's data into account, since there are significant randomness and the travel purposes are uncertain, and there are no clear passenger flow patterns, which have a significant influence on the learning process of the model and affect the prediction accuracy. Therefore, for further studies, researchers can overcome these deficiencies.

# Conflicts of Interest

The authors declare no conflict of interest.

# Acknowledgments

This work was supported by the Fundamental Research Funds for the Central Universities (No. 2021RC270), the National Natural Science Foundation of China (Nos. 71621001, 71825004), and the State Key Laboratory of Rail Traffic Control and Safety (Contract No. RCS2022ZQ001), Beijing Jiaotong University).